\documentclass[letterpaper, 10pt, conference]{ieeeconf}      

\usepackage{paralist}%
\usepackage{listings}
\usepackage{amsmath}
\usepackage{multirow}
\usepackage{graphicx}%
\usepackage{float}
\usepackage{listings}
\usepackage{caption}
\usepackage{subcaption}
\usepackage{hyperref}
\usepackage[flushleft]{threeparttable}

\newfloat{lstfloat}{htbp}{lop}
\floatname{lstfloat}{Listing}

\newcommand{\rcitem}[1]{\textsf{\small {#1}}} 
\usepackage{color}
\definecolor{dkgreen}{rgb}{0,0.6,0}
\definecolor{gray}{rgb}{0.5,0.5,0.5}
\definecolor{mauve}{rgb}{0.58,0,0.82}

\graphicspath{ {figures} } 

\IEEEoverridecommandlockouts                              

\usepackage{graphics} 

\title{%
  \LARGE \bf%
  Probabilistic modelling and safety assurance of \\ an agriculture robot providing light-treatment%
}%

\author{%
  Mustafa Adam$^{1}$, %
  Kangfeng Ye$^{2}$, %
  David A.\ Anisi$^{1,3}$, %
  Ana Cavalcanti$^{2}$, %
  Jim Woodcock$^{2}$, %
  and %
  Robert Morris$^{4}$
  \thanks{%
    $^{1}$ Robotics Group, Faculty of Science \& Technology, Norwegian University of Life Sciences (NMBU), Norway {\tt\small \{mustafa.adam, david.anisi\}@nmbu.no}}%
  \thanks{$^{2}$Dept.\ of Computer Science, University of York, UK {\tt\small \{kangfeng.ye, ana.cavalcanti, jim.woodcock\}@ york.ac.uk}}%
  \thanks{$^{3}$Dept.\ of Mechatronics, Faculty of Engineering and Science, University of Agder (UiA), Norway {\tt\small }}%
  \thanks{$^{4}$Saga Robotics AS, Norway
  {\tt\small rmorris@sagarobotics.com}}%
}%

\begin{document}

\maketitle


\thispagestyle{empty}
\pagestyle{empty}


\begin{abstract}
 Continued adoption of agricultural robots postulates the farmer's trust in the reliability, robustness and safety of the new technology. This motivates our work on safety assurance of agricultural robots, particularly their ability to detect, track and avoid obstacles and humans. This paper considers a probabilistic modelling and risk analysis framework for use in the early development 
 phases. 
 Starting off with hazard identification and a risk assessment matrix, the behaviour of the mobile robot platform, sensor and perception system, and any humans present are captured using three state machines. 
 An auto-generated probabilistic model is then solved and analysed using the probabilistic model checker PRISM. The result provides unique insight into fundamental development and engineering aspects by quantifying the effect of the risk mitigation actions and risk reduction associated with distinct design concepts. These include implications of adopting a higher performance and more expensive Object Detection System or opting for a more elaborate warning system to increase human awareness. Although this paper mainly focuses on the initial concept-development phase, the proposed safety-assurance framework can also be used during implementation, and subsequent deployment and operation phases.  
\end{abstract}

\section{Introduction}\label{sec:Introduction}

\begin{figure}[tbp!]
  \centering
  \includegraphics[width=0.9\linewidth]{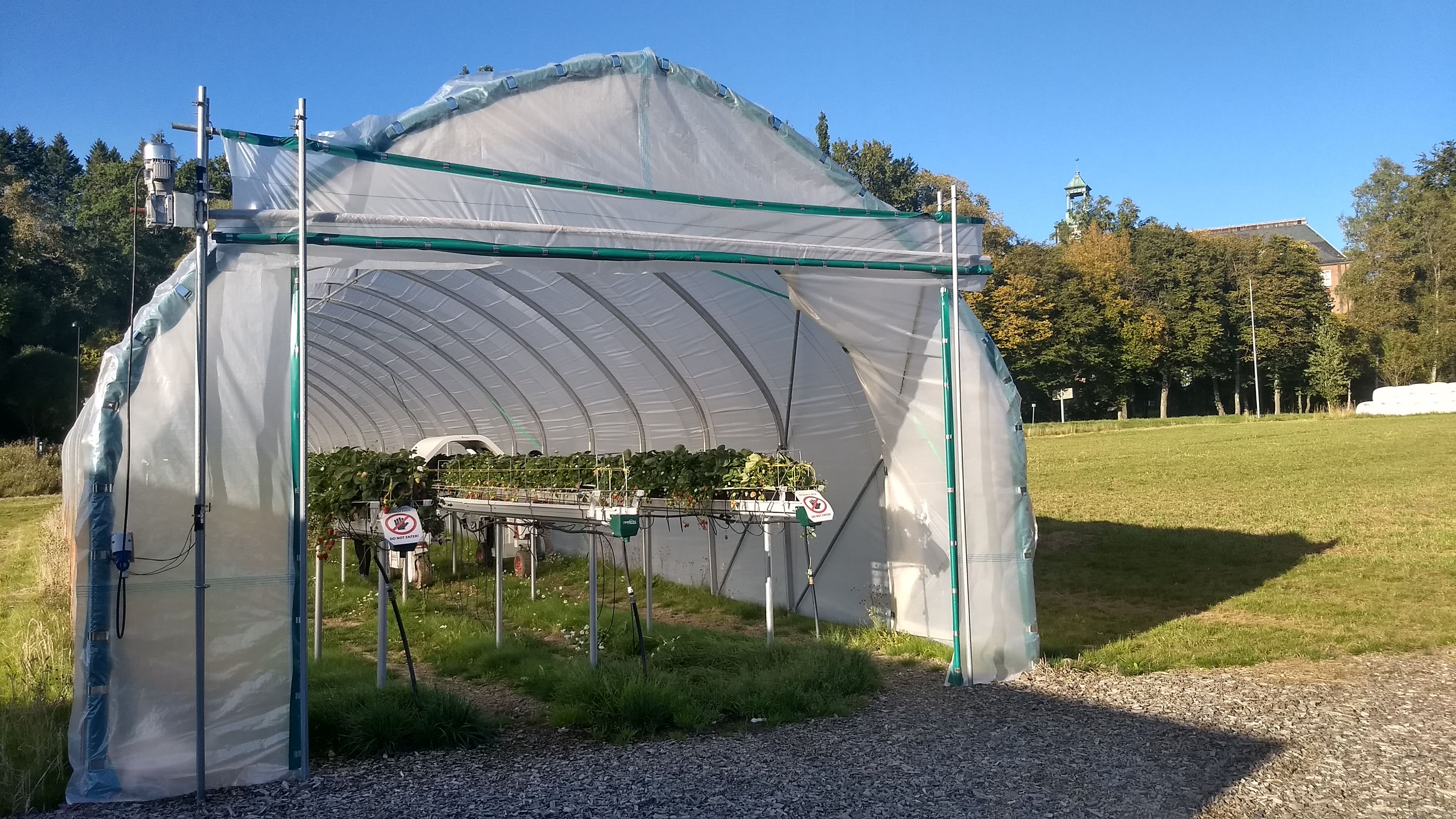}
  \caption{Robotic UVC light-treatment in poly-tunnel at NMBU}
  \label{fig:robot_treate_plants}
\end{figure}

The use of robots in agricultural tasks can:~(a)~improve efficiency and productivity, (b)~counter the shortage of seasonal workers, and (c)~cater for laborious and possibly dangerous tasks to protect humans from hazardous situations, such as spraying, mowing, pruning and light treatment as depicted in Fig.~\ref{fig:robot_treate_plants}. Technological advances in sensing, actuation, and machine learning have allowed more agricultural tasks to be carried out by  Robotic Autonomous Systems~(RAS).

One area of agriculture where RAS are especially useful is plant treatment. Many plant-pathogenic microorganisms, including fungi, damage the yield~\cite{doi:10.1094/PDIS-12-15-1440-RE}. Strawberries, particularly, are a high-value crop whose plants are subject to rapid degradation in taste and yield once affected. 
With a growing interest in reducing, or even forbidding, use of chemicals and fungicides, farmers are ready to consider alternatives. Research and real-life usage have shown that ultra-violet~(UV) radiation efficiently reduces disease development in many species, including strawberries~\cite {Impact_of_UV}. As UVC light is harmful to the human eye and is highly repetitive, and the light treatment is performed during the night and does not require  physical interaction with the 
plants, it is ideal for automation. UVC light-treatment services  to control powdery mildew using RAS are now commercially available. 

From historical data, it is evident that the agricultural domain is more willing and capable of adopting new solutions and technology as compared to other segments. This, however, postulates the farmer's trust in the new technology's safety~\cite{safety6010001}.
Therefore, assurance of agricultural robots is paramount to the continued adoption of novel robotic solutions. In this setting, the robot's ability to detect, track and avoid obstacles and humans is particularly interesting. 
 
Safety is defined as the absence of unacceptable risk, and relevant standards require systematic identification, measurement and monitoring of these risks~\cite{iso_agricultural_machinery,SafetyHandbook_smith2020}. To this end, any sound safety-engineering practice starts with the identification of all foreseeable \emph{hazards} that can cause harm to people, the environment or business. The associated \emph{risks} are then calculated by multiplying the severity of the hazards with their probability of occurrence. A \emph{risk-assessment matrix}, where the probability and severity of the hazards are depicted as rows and columns, is a fundamental tool in all hazard-based safety-engineering practices.

A highly relevant safety standard for RAS in the agricultural domain is ISO~18497~\cite{iso_agricultural_machinery}, while IEC~61508 is a cross-industry functional-safety standard~\cite{SafetyHandbook_smith2020}. Other standards, such as ISO~31000 and ISO~21000, indicate how to perform risk assessment and mitigation actions.  
Identifying and managing hazards and functional failures are 
some of the main challenges, particularly when involving AI or learning-based components~\cite{batarseh2021survey,SafetyHandbook_smith2020} Several approaches are often used, such as FTA (fault-tree analysis), FMEA (failure mode and effect analysis) and its variants, model-based reasoning, qualitative reasoning, and assumption-based truth maintenance~\cite{zaki2007detecting}.

In extension to the prior art, this paper presents 
an approach for analysing the risks associated with UVC treatment during row transition. Starting with the hazard identification results in~\cite{leo_MeSAPro_report}, we use a diagrammatic domain-specific notation, namely, RoboChart~\cite{RoboChartSoSym}, to model the behaviour of the mobile robot platform, sensor, and perception system, as well as humans. Using RoboChart, we define three synchronised probabilistic state machines. From a RoboChart model, a probabilistic model can be auto-generated, then solved and analysed using the PRISM model checker~\cite{Ye2022}. Here, based on this, risk mitigation plans and several design concepts are proposed to control risk. In particular, results regarding the effect of ODS on reducing the risk below tolerable level are presented. In~\cite{RoboChartSoSym,Ye2022}, RoboChart is used to model existing RAS, including the robot platform, controller and mechanical components. This paper complements that work by utilising the same modelling and reasoning framework to capture and quantify risks during the early development phase. 

The remainder of this paper is organised as follows. Next, we discuss related work. Section~\ref{sec:case_study} describes the light-treatment use case.  Section~\ref{sec:probabilistic_model_check} describes the adopted probabilistic model-checking methodology. 
Section~\ref{sec:fsm} explains the RoboChart models and the safety-properties formalisation. The analysis results are in Section~\ref{sec:results}. Finally, conclusions and directions for future work are in Section~\ref{sec:conclusion}.

\section{Related work}\label{sec:related-work}

Mayoral \emph{et al.}~\cite{jose_navigation,jose_risk_management_based} also considers the safety of humans near agricultural robots. The sensor is an RGBD camera, and a neural network architecture~(YOLOv4) is used to classify humans concerning their distance from the robot. This complementary work sets the stage for the Object Detection System (ODS) considered in this paper.  

\begin{figure}[tbp!]
  \centering
  \includegraphics[width=0.9\linewidth]{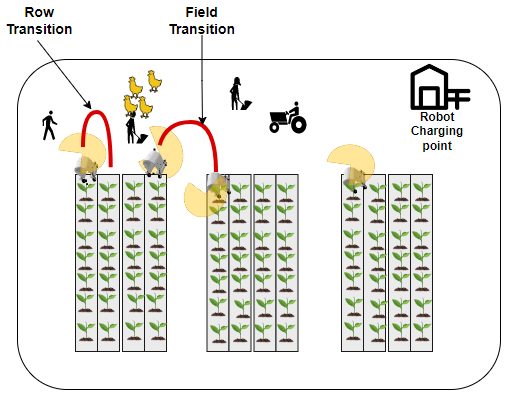}
  \caption{Robot(s) performs farming operations. The robot can perform different tasks along the rows and then do a transition at the end of row. During rows/fields transition, it might encounter other living or obstacles. We are interested in row transition risk during UVC treatment. }
  \label{fig:field-transition}
\end{figure}

Regarding the robotic light treatment of strawberries, \cite{leo_MeSAPro_report} presents a list of potential hazards and failure modes during UVC treatment identified using FMEA. Table~\ref{table:hazards_cases} gives one example. There, the probability of a human getting injured by the UVC light when varying the occurrence of the failures is quantified. The results indicate that the riskiest scenario occurs when the robot is unaware of a human approaching from the side while transitioning between rows~(Fig.~\ref{fig:field-transition}).

Based on~\cite{leo_MeSAPro_report,Impact_of_UV}, we study here the probability of the robot being unaware of the presence of a human during row transition until it is too late to act. 
Our focus is on the risk of human-robot encounters during row transition (F-G5 in Table~\ref{table:hazards_cases}), but the fundamentals apply to other hazards and scenarios. For the current example, the field transition, in which the robot moves between different parts of the field (see Fig.~\ref{fig:field-transition}), also presents risks we can address.

\begin{table*}[hbtp!]
  \centering
  \caption{Hazard identification during the studied scenario, from~\cite{leo_MeSAPro_report}.}
  \begin{threeparttable}
  \begin{tabular}{p{0.2\linewidth}p{0.05\linewidth}p{0.1\linewidth}p{0.1\linewidth}p{0.1\linewidth}p{0.06\linewidth}p{0.06\linewidth}}  
  \hline    
    Possible situation & Code & Possible failures & Potential effect & Consequence & Severity & Occurrence\\
    \hline
    \multirow[c]{2}{3cm}{Robot at the end of the rows when a worker is approaching laterally} 
    & F-G5
    & Robot detects the human only when they are too close (less than 3.6 m)
    & Robot stop using UV-C light too late
    & Human is getting injured by the UV-C light
    & critical
    & probable \\
    \hline
  \end{tabular}   
 
  \end{threeparttable}
 \label{table:hazards_cases}  
\end{table*}

\section{Light-Treatment Use-Case}  \label{sec:case_study}

The UVC treatment is a preventive measure to be applied weekly to control powdery mildew. With this frequency, there is a non-negligible risk of possible robot encounter with field workers, untrained people and visitors, kids, and animals~\cite{XIE2016337}. A mitigation plan to reduce risks should consider the distance from the lamp and exposure time~\cite{PMID:23722672}. 

\subsection{Robot protection system}

\begin{figure}[tbp!]
  \centering
  \includegraphics[width=0.5\linewidth]{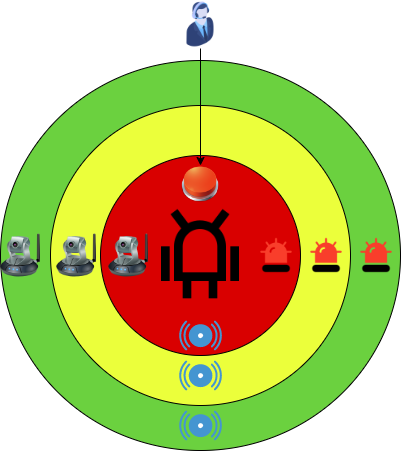}
  \caption{Multi-layered protection system. To minimize human injuries, multi-layered protection system is deployed. Anomalies can be detected by sensors, camera and LiDar readings on different distances (green-yellow-red regions). Warning sub-system and emergency stop button are also included}
  \label{fig:multi_layered_protection}
\end{figure}

Fig.~\ref{fig:multi_layered_protection} demonstrates the robot protection layers, which are activated based on the distance of a human from the robot. Hazard zones are defined as follows. The green zone is the area more than 7~m away from the robot. It is outside the influence of the system but can be monitored to assess the potential for interaction. The yellow area is where a person may be approaching and can be monitored by the system; it is estimated to be between 3 and 7~m from the robot. In the red zone, harm may occur:~0-3~m from the robot~\cite{leo_MeSAPro_report}.

The robot's safety system has three main components. The primary system uses visual and audio warnings. The secondary sensor-based system includes ODS, which can be one or a combination of distinct sensor types, most prominently 3D camera and LiDAR, as these are typically also utilised for navigation.
The final safety system includes impact-recognition bumpers and emergency stop buttons.

\subsection{Operational assumptions and safety property}

Here, we present a way to quantify the probability of human injury, or more precisely, of humans being at high risk inside the red zone during UVC treatment. Afterwards, we discuss mitigation plans to reduce the risk factors. The following assumptions have been considered to define the constants and probabilities used in the RoboChart model presented in Section~\ref{sec:fsm}. First, any human entering the robot's operational area can move in an arbitrary direction. The robot does not carry any  physical barrier that would prevent humans from entering the red zone. Second, the increased safety implied by the existing primary protection system is incorporated and accounted for in the approaching decision probabilities into the different zones. Third, our analysis focuses on ODS based on sensor readings that allow the detection of human presence within a distance of over~7m.  Finally, the damage is considered to have occurred once the human enters the red zone. This situation causes human injury and is used to define the safety property in Section~\ref{sec:fsm}. 

According to the hazard identification  in~\cite{leo_MeSAPro_report}, we notice a major risk associated with operation O-U1, failure F-G5. We aim to reduce this risk by reducing its occurrence probability.  
Utilising the ODS as a Safety Instrumented System~(SIS) and reducing the maximum tolerable occurrence probability, a higher Safety Integrity Level~(SIL) can be obtained~\cite{SafetyHandbook_smith2020,sil}. We can reduce the risk associated with F-G5 by an order of magnitude so that it reaches tolerable levels. 

\section{Probabilistic model checking using RoboTool and PRISM}  \label{sec:probabilistic_model_check}

Formal verification ensures that a system fulfils given specifications in all circumstances regardless of input possibilities~\cite{DBLP:journals/corr/abs-2012-00856,Peled2019,8675610}. In our use case, machine learning components are enabled in the robot. Due to the probabilistic nature of such components, probabilistic model checking is used to capture the behaviour of the perception system and its interaction with the environment~\cite{PRISM_2009,probabilistic_modelling}.

The UVC treatment and hazard scenario F-G5 from Table~\ref{table:hazards_cases} is modelled as a Discrete-Time Markov Chain (DTMC). The transition between one state to another is assumed to be deterministic and depends on event probabilities. 
 
RoboChart is a domain-specific language for model-based robotics software engineering, with formal semantics encompassing functional, timed, and probabilistic aspects tailored for formal verification. Its tool is called RoboTool\footnote{\textrm{robostar.cs.york.ac.uk/robotool/}}~\cite{RoboTool,RoboStarTechnology2021}. Capturing the UVC-treatment behaviour in an abstract and formal modelling notation like RoboChart can be challenging, especially for practitioners who are used to working with code-driven, dynamic simulations based on sketches of system decomposition and design. This can be mitigated by training, and capturing the behaviours in PRISM directly is even more challenging for practitioners without knowledge of formal methods and PRISM. In our use case, for example, it is difficult to model the naturally asynchronous behaviour of the human, robot and ODS in terms of a global clock in PRISM. Here, asynchronous behaviour means they progress at their own pace but only at a time tick. In addition, RoboChart, through its diagrammatic notation, helps a developer to focus on the high-level design of the application, to write a correct model, instead of on analysis details. This can reduce verification cycles and improve efficiency.

A model checker requires use of bounded data types; it goes through all possible states and transitions to verify or disprove a property. Specific values for variables or inputs can be used, but that restricts the range of values analysed. Thus, keeping the right level of abstraction during modelling is essential for capturing all relevant aspects of the system, getting a meaningful result from model checking, and keeping the computational complexity at bay.

This paper uses RoboTool to automatically generate a PRISM model from our RoboChart model, and a PRISM property file from properties described using RoboChart's  controlled natural language~\cite{Ye2022}. RoboTool runs multiple PRISM instances~(229 for this use case and one for each property) in the background to analyse properties simultaneously. This procedure is fully automated. The generated PRISM model is larger than initially expected, as usual for automatically generated models. It captures not only the architecture of the RoboChart model, but also its complete semantics, such as all interactions between controllers and state machines, composite states, high-level transitions and actions in a low-level command-based PRISM language. 

The two properties to be verified are formalised using the probabilistic temporal logic PCTL~\cite{WAN2013279}. With RoboTool, we do not need to use PCTL to describe the properties:~a controlled natural language is available, from which PCTL formulas can be automatically generated. 
In this section, we define the properties in PCTL and present the RoboTool facility in the next section. We define the property that captures the probability of human injury during UVC treatment on row transition using the operator \textit{P}. It is used in a quantitative approach to reason about the probability of event occurrence in $P_{=?} [path~property]$. The formalisation is as follows:

\begin{enumerate}
 \item[\textbf{Property P1 (Injury):}]\label{Prop_P1} 
\begin{equation}
 \begin{split}
P_{=?} \left[F \left(
    \begin{array}{l}
shuman = inRed ~\land \\
srobot = transitionRow ~\land \\
ticks = t
\end{array}\right)\right]  
\end{split}
\end{equation}
This is a query of the probability ($P_{=?}$) of the system finally ($F$) reaching a situation where the human is in the red zone~($shuman = inRed$), the robot is in the transition row ($srobot = transitionRow$), and the number of $ticks$ is that of a parameter $t$ of the query.
\end{enumerate}


\noindent%
The following property checks deadlock freedom.
\begin{enumerate}
 \item[\textbf{Property P2 (Deadlock):}]\label{Prop_P2} 
\begin{equation}
 \begin{split}
\lnot E~[F~\text{``deadlock''}]  
\end{split}
\end{equation}
This property requires that does not ($\lnot$) exist ($E$) a path such that finally ($F$) the system deadlocks. Here, ``deadlock'' is a predefined label that identifies the states without outgoing transitions.
\end{enumerate}

\noindent%
Next, we present our RoboChart models and properties as modelled using controlled natural language. 

\section{State Machines and Property Modelling} 
\label{sec:fsm}

\begin{figure}[tbp!]
    \centering
    \includegraphics[width=0.95\linewidth]{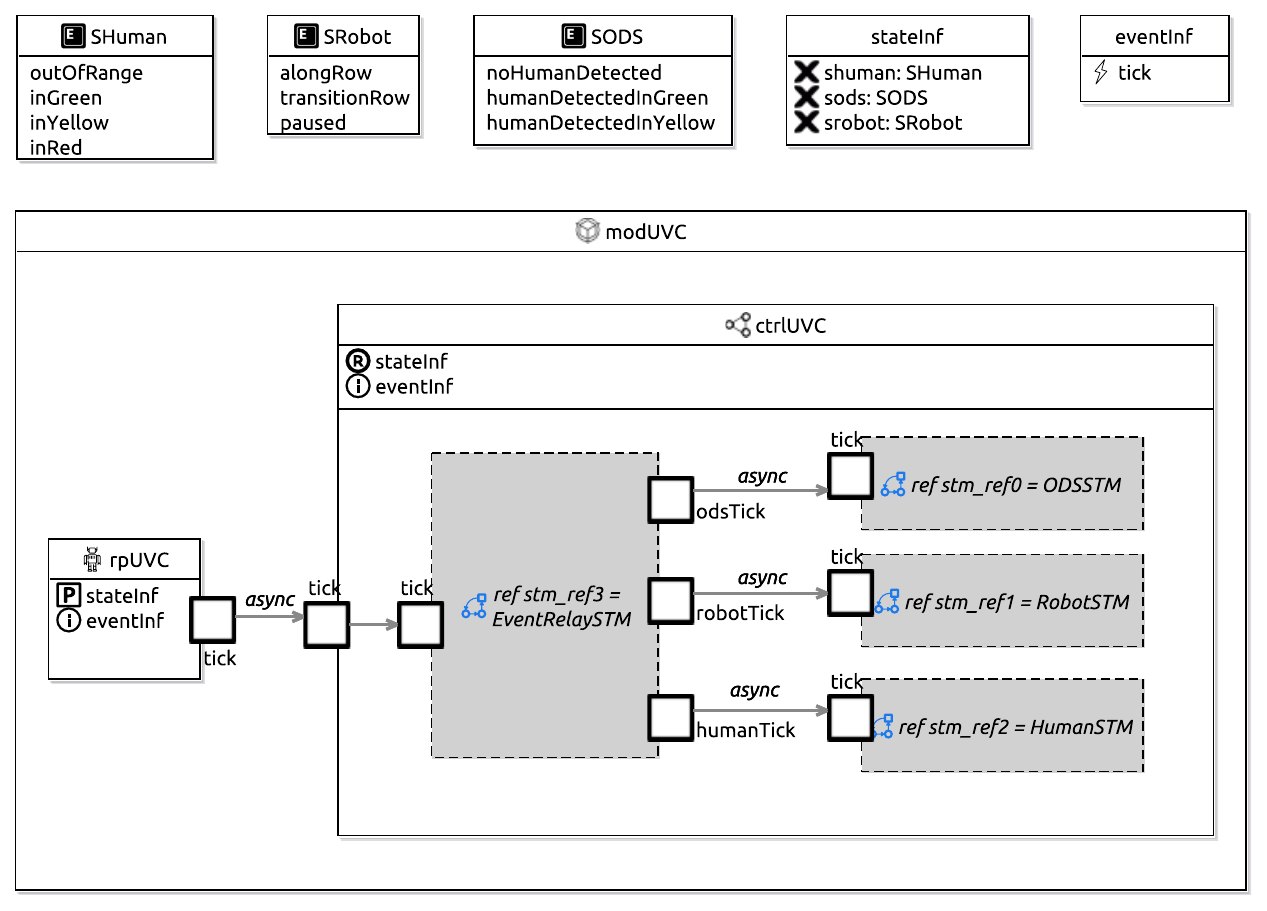}
    \caption{UVC RoboChart component model. The controller receives clock-generated ticks. These ticks are distributed to three state machines:~\rcitem{ODSTM}, \rcitem{RobotSTM}, and \rcitem{HumanSTM}.}
    \label{fig:uvc_module}
\end{figure}

In Fig. \ref{fig:uvc_module}, we show a RoboChart component model~(in a block \rcitem{modUVC}) describing the high-level architecture of the whole system. The use case implementation is available on RoboStar technology GitHub. \footnote{\url{https://github.com/UoY-RoboStar/uvc-case-study/}} In that model, 
a robotic platform block \rcitem{rpUVC} describes an abstraction of the robot via three shared variables:~\rcitem{shuman}, \rcitem{sods}, and \rcitem{srobot}, declared in the interface block \rcitem{stateInf} to record the current status of each entity. In another interface \rcitem{eventInf}, we define an event \rcitem{tick} to synchronise the behaviour of the entities. 
Inside \rcitem{modUVC}, a controller block called \rcitem{ctrlUVC} contains four state machines:~\rcitem{ODSSTM}, \rcitem{RobotSTM}, and \rcitem{HumanSTM}, to specify the behaviours of the entities, and \rcitem{EventRelaySTM}, relaying the \rcitem{tick} event from the platform to \rcitem{EventRelaySTM}.
    
\begin{figure}[tbp!]
    \centering
    \includegraphics[width=0.95\linewidth]{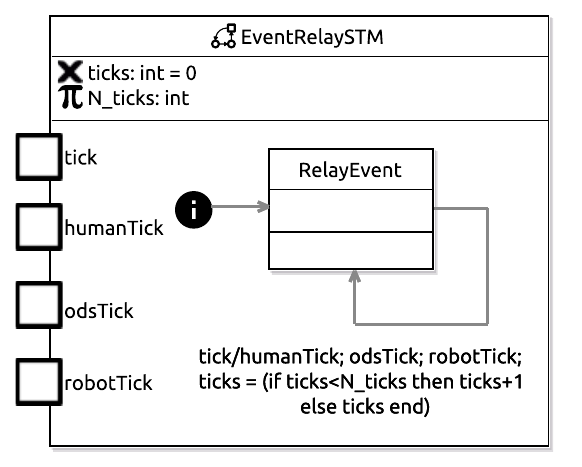}
    \caption{UVC RoboChart model: event relay state machine. Keeps generating tick messages until N\_ticks is reached.}
    \label{fig:uvc_event}
\end{figure}

\begin{figure}[tbp!]
    \centering
    \includegraphics[width=0.95\linewidth]{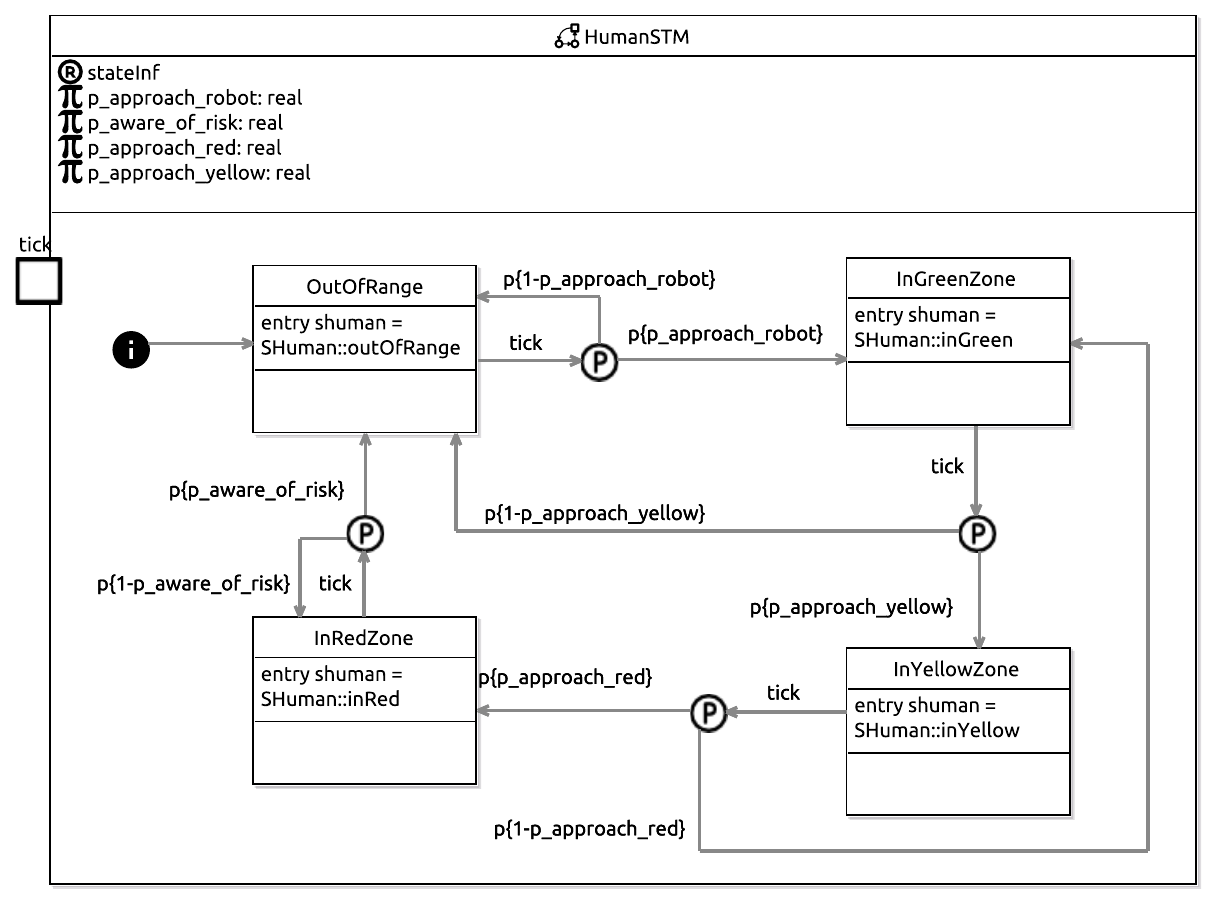}
    \caption{UVC RoboChart model: human state machine.}
    \label{fig:uvc_human}
\end{figure}

\begin{figure}[tbp!]
    \centering
    \includegraphics[width=0.9\linewidth]{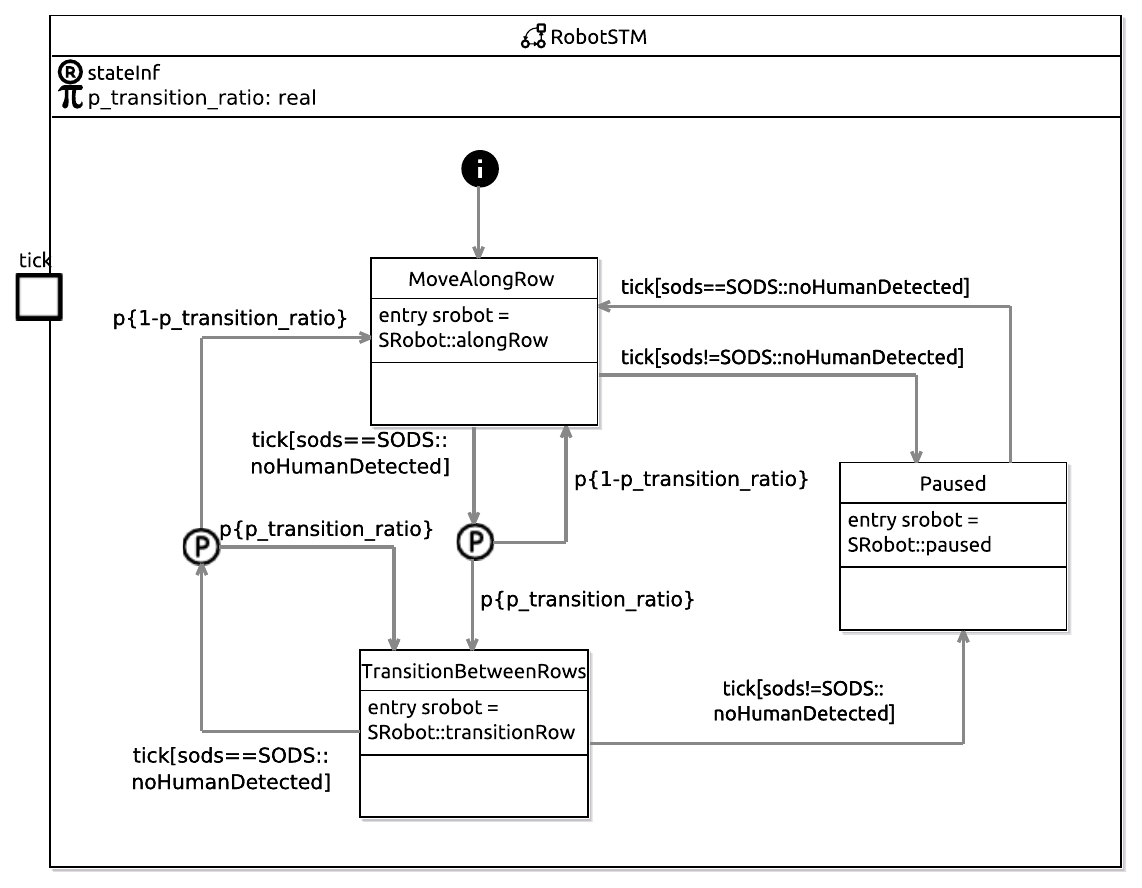}
    \caption{UVC RoboChart model: robot state machine.}
    \label{fig:uvc_robot}
\end{figure}

\begin{figure}[tbp!]
    \centering
    \includegraphics[width=0.9\linewidth]{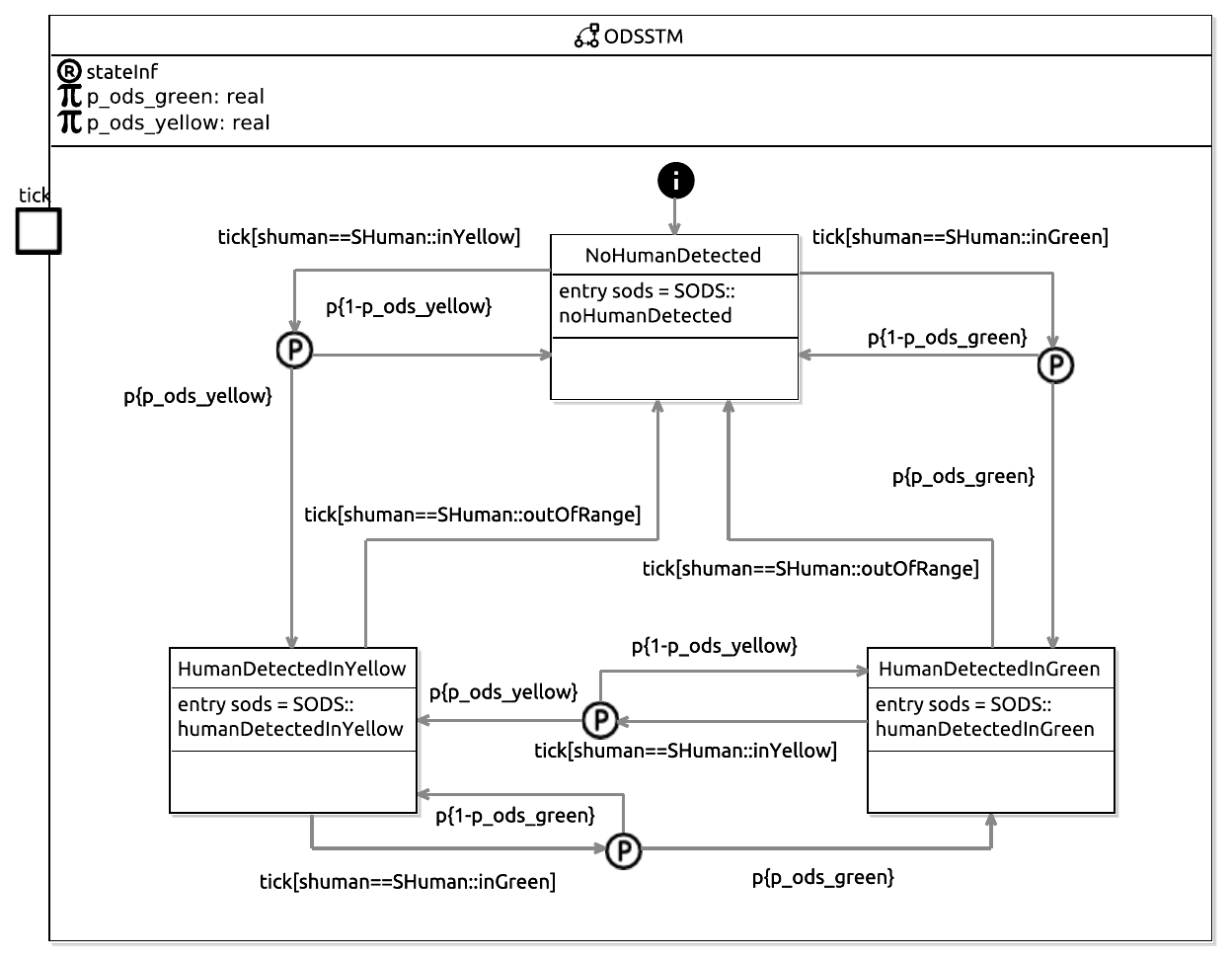}
    \caption{UVC RoboChart model: ODS state machine.}
    \label{fig:uvc_ods}
\end{figure}

The connections between the platform and \rcitem{EventRelaySTM}, and between \rcitem{EventRelaySTM} and the other three state machines are \rcitem{async}hronous:~a one-place buffer in RoboChart. Every \rcitem{tick} from the platform adds to or overrides the buffer connected to \rcitem{EventRelaySTM}. As shown in Figs.~\ref{fig:uvc_event}-\ref{fig:uvc_ods}, the machines need to take a \rcitem{tick} event from the buffer for a transition to be taken. After the event is taken, the buffer is empty, and the state machine has to wait till the next \rcitem{tick} to progress. With the \rcitem{EventRelaySTM} machine, each \rcitem{tick} is passed on from the platform to the other machines. We believe this protocol is the most difficult to be implemented in PRISM directly without using RoboChart.

The variable \rcitem{ticks} records the number of steps of the system~(akin to the passage of global time).  With that, we can formalise our property of interest, namely, the probability of a human entering a dangerous zone at different steps. With the constant \rcitem{N\_ticks}, we can bound the value of \rcitem{ticks} as required for model checking~(in PRISM). The value of \rcitem{N\_ticks} is fixed during translation to PRISM so that our verification can explore different values.

The machines \rcitem{HumanSTM}, \rcitem{RobotSTM}, and \rcitem{ODSSTM} start at the state targeted by the transition from their initial junctions~(a black circle with an \rcitem{i}). For example, \rcitem{HumanSTM} in Fig.~\ref{fig:uvc_human} enters the \rcitem{OutOfRange} state, which captures behaviour when the human is out of the robot operation area. \rcitem{InGreenZone}, \rcitem{InYellowZone}, \rcitem{InRedZone} are concerned with behaviour when the human's position relative to the robot is in each of the zones in Fig.~\ref{fig:multi_layered_protection}. The transitions from one state to another have a trigger \rcitem{tick} and go via a probabilistic junction where a choice is made based on a configurable probability defined by a constant whose value is left open. As explained later in Section~\ref{section:constants}, each probability is based on awareness of the risk to get closer or leave the zone. 

\rcitem{RobotSTM} in Fig.~\ref{fig:uvc_robot} starts 
at the state \rcitem{MoveAlongRow}, capturing UVC treatment along the row, and alternates between \rcitem{MoveAlongRow} and \rcitem{TransitionBetweenRows} with a ratio \rcitem{p\_transition\_ratio} such as 10:1. This effectively implies that the robot spends ten-fold time in \rcitem{MoveAlongRow} as compared to \rcitem{TransitionBetweenRows}. If a human is detected in green or yellow zones, the robot will move to \rcitem{Paused} via the transition with guard \rcitem{[sods!= noHumanDetected]}. 

In Fig.~\ref{fig:uvc_ods}, \rcitem{SODS} enters \rcitem{NoHumanDetected} upon initialisation. Afterwards, human position and system detection accuracy dictate the next state. The system then can move to \rcitem{HumanDetectedInGreen} with accuracy \rcitem{p\_ods\_green} or \rcitem{HumanDetectedInYellow} with accuracy \rcitem{p\_ods\_yellow}. 

In the next section, we study the values of the probabilities, and in Section~\ref{sec:safety_property}, we formalise safety.

\subsection{Probabilities and constants definitions}
\label{section:constants}

    A human decides to get close to the robot with three different probabilities:~\rcitem{p\_approach\_robot}, \rcitem{p\_approach\_yellow}, and \rcitem{p\_approach\_red}. The decision is based on awareness and previous training. We differentiate between three levels of awareness as described in  Table~\ref{table:awarness_levels}. Finally, we set \rcitem{p\_aware\_of\_risk} to 0.01, since, as already said, the damage already occurs when a person is in the red zone.
    
    The \emph{ODS} accuracy is characterised by  two probabilities:~\rcitem{p\_ods\_green} is the accuracy in detecting an object within the green zone, and \rcitem{p\_ods\_yellow} in the yellow zone. In the study, we set a higher value for \rcitem{p\_ods\_yellow} since detecting an object closer to the sensor is easier. 
    
    The scenarios considered are as follows. 
    (1)~High performance ODS system:~both \rcitem{p\_ods\_yellow} and \rcitem{p\_ods\_green} are  \textrm{0.99}. 
    (2) Normal ODS:~\rcitem{p\_ods\_yellow} and \rcitem{p\_ods\_green} are set to \textrm{0.7} and \textrm{0.4}.
    (3)~Non-functioning or lacking ODS system:~\rcitem{p\_ods\_green} and \rcitem{p\_ods\_yellow} are set to \textrm{0}.

\begin{table}[tbp!]
  \centering
  \caption{Awareness levels in the experiment}
  \begin{tabular}{|p{1.2cm}|p{3cm}|p{3cm}|}  
  \hline    
    Level & Description & Probabilities\\ \hline
    Deliberate
    & A person is determined to reach the robot. For instance, if curious and unaware of the risk.
    &  \textrm{p\_approach\_robot=1} \textrm {p\_approach\_yellow=1}
     \textrm{p\_approach\_red=1} \\ \hline
    Aware
    & Even if the person has a higher chance of entering the green and yellow zones, entering the red zone is less probable.
    &  \textrm{p\_approach\_robot=0.5} \textrm{p\_approach\_yellow=0.5}
     \textrm{p\_approach\_red=0.3} \\ \hline
    
    Less Aware
    & The person is unaware of the risk but cautious about entering the red zone.
    &  \textrm{p\_approach\_robot=0.7}
     \textrm {p\_approach\_yellow=0.7}
     \textrm{p\_approach\_red=0.5} \\\hline
  \end{tabular}
  \label{table:awarness_levels}  
\end{table}

\subsection{Formulation of the safety property}
\label{sec:safety_property}


\noindent
As already mentioned, to specify properties and assertions, such as Properties~\textbf{P1} and~\textbf{P2} in section ~\ref{sec:probabilistic_model_check}, RoboTool provides a simple and more readable textual domain-specific language. Listing~\ref{listing:robocert} presents the safety property {\bf P1}. Values of constants and probabilities are picked up from definitions in a configuration named \texttt{C1}. A configuration is just a list of constant names and their associated values, that we need to define in RoboTool to support proof my model checking. Listing ~\ref{listing:deadlock} defines the deadlock property \textbf{P2}. 

\begin{lstfloat}
\begin{lstlisting}
import uvc_config::*

prob property P1:
  Prob=? of [Finally 
    modUVC::rpUVC::shuman==inRed /\ 
    modUVC::rpUVC::srobot==transitionRow /\
    modUVC::ctrlUVC::stm_ref3::uvs==t]
  with constants C1
	
\end{lstlisting}
\caption{\label{listing:robocert}Property definitions in RoboTool of human injury. C1 denotes probability values in the configuration file.} 
\end{lstfloat}

\begin{lstfloat}
\begin{lstlisting}
prob property P2:
	not Exists [Finally deadlock]
	with constant C1
	
\end{lstlisting}
\caption{\label{listing:deadlock} Deadlock property definition}
\end{lstfloat}

\section{Model-Checking Results} \label{sec:results}

\begin{figure}[tp!]
\begin{subfigure}[b]{0.5\textwidth}
    \centering
    \includegraphics[scale=0.55]{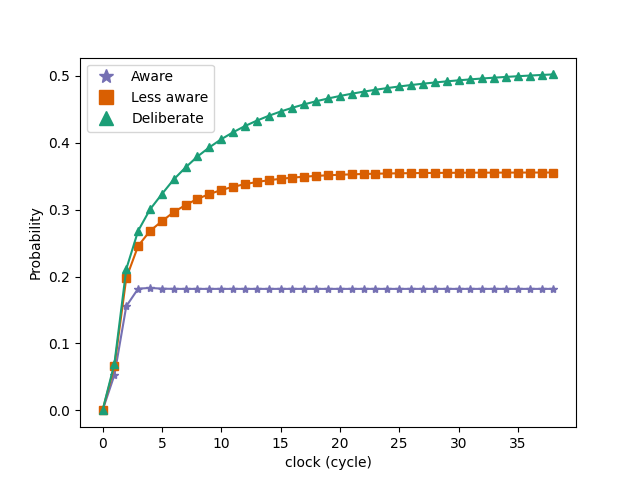}
    \caption{ODS malfunctioning: Failure}
    \label{fig:no_ods}
\end{subfigure}
     \hfill
\begin{subfigure}[b]{0.5\textwidth}
    \centering
    \includegraphics[scale=0.55]{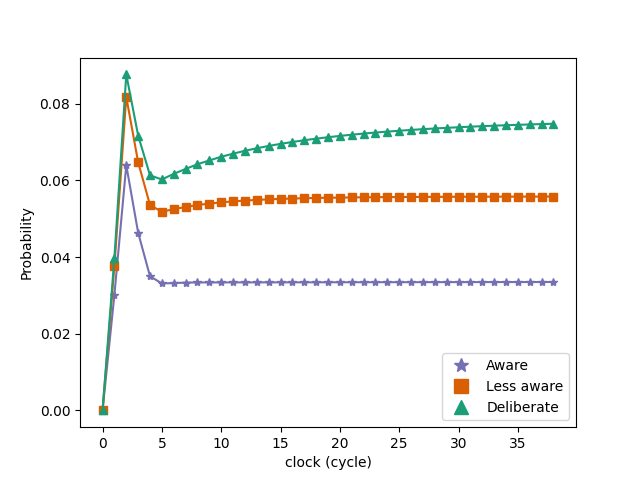}
    \caption{Normal performance ODS}
    \label{fig:normal_ods}
\end{subfigure}
     \hfill
\begin{subfigure}[b]{0.5\textwidth}
    \centering
    \includegraphics[scale=0.55]{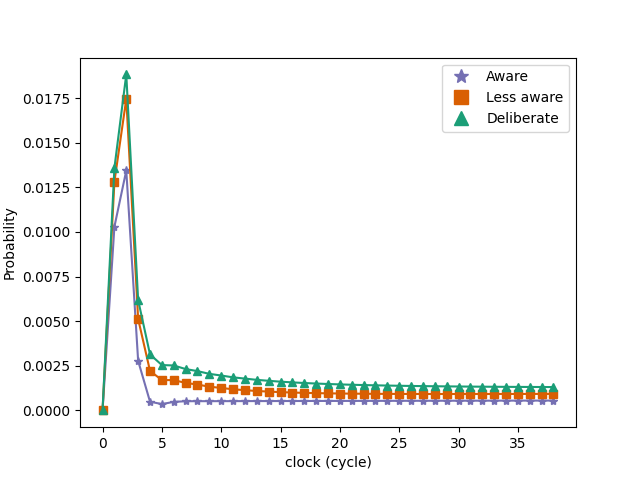}
    \caption{High performance ODS}
    \label{fig:hp_ods}
\end{subfigure}
\hfill

\caption{Probability of human injury when encountering the robot during row transition for various ODS safety systems and human awareness levels.}
\label{fig:three graphs}
\end{figure}

During the experiments based on Property~\textbf{P1} using PRISM, the level of awareness and type of ODS system are changed. 
 As expected and evident in all graphs depicted in Fig.~\ref{fig:three graphs}, the probability of injury during row transition is reduced when the awareness level of the human is increased.

\begin{figure}[htbp!]
    \centering
    \includegraphics[width=\linewidth]{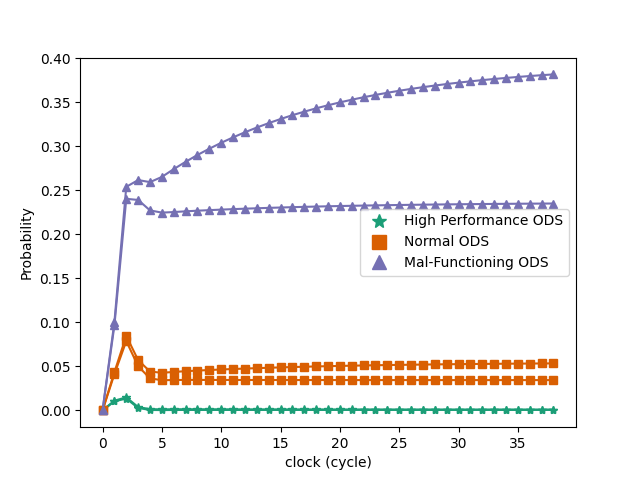}
    \caption{Comparison between different ODS systems when a person is aware and  deliberately approaching the robot. The risk is reduced by an order of a magnitude when high-performance ODS is applied while awareness does not reduce risk level to an acceptable level}
    \label{fig:combined}
\end{figure}

Next, attention is turned towards the effect of the ODS on the occurrence probability of hazard~F-G5 as elaborated upon in Section~\ref{sec:Introduction}. Utilising the ODS for safety purposes helps reduce the occurrence probability and enables us to push the associated risk below an acceptable level.  As F-G5 leads to human injury with high likelihood and thus the risk of this hazard is high~\cite{leo_MeSAPro_report}, the maximum tolerable occurrence probability must be sufficiently low. 

From Fig.~\ref{fig:no_ods} and Fig~\ref{fig:normal_ods}, it is concluded that the probabilities of injury are above \textrm{0.1} and \textrm{0.01} when the ODS is not functioning or using an average-performing ODS. 
Fig.~\ref{fig:combined} compares the effect of hazard occurrence probability as a function of ODS type. This figure presents results considering both deliberate and aware human behaviour and depicts how accurate and high-performance sensors reduce hazard occurrence probability. Compared to not using any ODS or having a malfunctioning one, the normal ODS with average quality provide a risk reduction factor 10. This represents a Safety Integrity Level (SIL) of 1. Opting for a high-performance ODS yields a risk reduction factor of 100, representing SIL 2. 
These results emphasise the importance of improving the ODS system when designing a safety architecture for agriculture RAS~\cite{jose_navigation}.  

To increase the confidence in the correctness of the model, it was verified  that the UVC state machine could not deadlock; that is, Property~\textbf{P2} is also fulfilled.

\section{Concluding Remarks and Future Work}
\label{sec:conclusion}

Our main contributions are a risk analysis and safety assurance approach for agricultural robots. We use probabilistic model checking to quantify the risk of human injury due to UVC-light exposure. 
The results give insight and guidelines during the early development phase on improving the safety system and implementing a risk mitigation plan. These include implications of improving the detection algorithm, adopting a higher performance and more expensive sensors, or improving the safety policies through more elaborate warning systems to increase human awareness.

As Functional Safety Assessment is needed in all phases of development, a natural progression of the current work includes:~%
(1)~engineering and realisation of (high-performance) ODS including machine-learning components~\cite{jose_navigation};
(2)~development and formal verification of navigation and control laws for the agricultural robot platform, taking both hardware and software components into  account explicitly~(co-verification)~\cite{murray2022safety};
(3)~formal verification and validation of ODS performance during the operation phase; and 
(4)~run-time monitoring and verification of Property~\textbf{P1}.  


\addtolength{\textheight}{-9cm}   
\section*{Acknowledgements}

The authors would  like to gratefully acknowledge all guidance and fruitful discussions provided by Robert Morris on the hazard and safety aspects of agricultural robots.   
The research presented in this paper has received partial funding from the Norwegian Research Council (RCN) RoboFarmer, project number 336712, the UK EPSRC Grants EP/M025756/1, EP/R025479/1, and EP/V026801/2, and  the Royal Academy of Engineering Grant No CiET1718/45.


\bibliographystyle{IEEEtran}
\bibliography{IEEEabrv,bibliography}

\end{document}